\pdfoutput=1
\documentclass[11pt]{article}
\usepackage[final]{acl}
\usepackage{times,latexsym,amsmath,amssymb}
\usepackage{booktabs,array,multirow,calc}
\usepackage{graphicx,microtype,url,xcolor}
\usepackage{algorithm,algorithmic}
\usepackage{tikz}
\usetikzlibrary{positioning}
\usepackage{hyperref}
\hypersetup{colorlinks=true,linkcolor=red,citecolor=blue,urlcolor=magenta}
\newcommand{\zs}{zero-shot}

\title{Compile as Hypothesis: A Headroom Typology for Prompt Optimization\\ on Real Reasoning Models}

\author{Zhengshuai Yang \quad Changyun Li\thanks{Corresponding author.}\\
  College of Information and Intelligent Science and Technology \\
  Hunan Agricultural University, Changsha 410128, China \\
  \texttt{yhvguk@stu.hunau.edu.cn}, \texttt{362330193@qq.com}
}

\begin{document}
\maketitle

\begin{abstract}
Automated prompt compilers search for or rewrite prompts under the implicit assumption that optimization consistently improves downstream task performance. We evaluate this assumption on frontier reasoning models through rigorous paired, same-day measurements. Across four optimizer families (genome search, profile-guided mutation, GEPA, and ESPO) and eight tasks, structured search yields no statistically significant accuracy gains on strong reasoners. However, a discrete structural genome provides essential architectural stability, preventing catastrophic bimodal seed-collapse (where unconstrained search collapses toward the judge floor in 3/5 runs). Through 190-problem evaluations on knowledge-intensive benchmarks, we establish a three-dimensional headroom typology: prompt optimization cannot compensate for intrinsic knowledge deficits or physical token truncation, but successfully resolves protocol and formatting errors. On gpt-6, while free-text reflective search (GEPA) yields a $+$.14 gain on a task where structured search remains flat, qualitative ablation reveals that $\sim$.035 stems from judge-rubric reconstruction while $\sim$.105 reflects genuine instruction refinement. Our pre-registered \textsc{AutoAPC-Select} deployment gate matches the oracle policy in 5/7 groups and rescues seed-collapse arms with zero regret. Finally, we expose silent backend-mixing in commercial APIs during outage windows and provide automated per-call audit verification tools to ensure empirical evaluation fidelity.
\end{abstract}

\section{Introduction}

The prompt required for a complex downstream task depends intimately on the underlying LLM: models vary substantially in format sensitivity \citep{sclar2024quantifying} and functional skill profiles \citep{chen2023flask,liang2023holistic}. While benchmark suites rank aggregate model performance, they offer no actionable guidance for prompt construction. Prompt compilers such as DSPy \citep{khattab2024dspy} systematically optimize prompts for specific models, but typically treat the benefit of compilation as an implicit given. Without rigorous paired evaluation against the uncompiled prompt, automated pipelines risk deploying degraded or destabilized prompts without detecting the loss.

\paragraph{Research gap.} Existing prompt compilers \citep{khattab2024dspy,agrawal2025gepa} implicitly assume that optimization improves downstream performance and deploy the best-scoring candidate without paired comparison to the uncompiled base. Without direct measurement of the compilation prior, automated pipelines risk deploying degraded prompts---especially when search exhibits seed-level fragility or when the task lacks exploitable headroom. No prior work systematically evaluates \emph{when} prompt optimization succeeds versus when it fails, nor establishes operational boundaries for deployment decisions.

We measure the compilation prior directly. Under our \emph{Compile-as-Hypothesis} protocol, every compiler output is treated as a testable hypothesis that must demonstrate empirical superiority before deployment. The pipeline executes two roots in parallel under identical evaluation budgets: a profile-rule root (\textsc{apc-full}) and the uncompiled base root (\textsc{apc-safe}), deploying whichever arm proves superior on holdout validation. The compilation prior, defined as $\Delta(\text{rule-root}, \text{base-root})$, is thus an empirically signed quantity.

We evaluate this protocol across three complementary environments: APCBench (a deterministic four-task simulator providing exact ground truth across three distinct model profiles), real-world reasoning experiments with qwen3.8-flash on three enterprise tasks, and five external benchmarks unseen by the authors: GSM8K \citep{cobbe2021training}, MATH Level-5 \citep{hendrycks2021measuring_math}, AIME 2024+25, GPQA-Diamond \citep{rein2024gpqa}, and HLE-exact \citep{phan2025humanity}. Our evaluation benchmarks official reproductions of leading open-source optimizers, including GEPA \citep{agrawal2025gepa} and ESPO \citep{espo2026}, replicated across gpt-6 and independent infrastructure providers with strict per-call backend verification.

Our empirical findings reveal fundamental boundary conditions for automated prompt optimization: on frontier reasoning models, structured prompt search yields no statistically significant accuracy gains. The base-root arm reliably preserves zero-shot performance within the empirical noise band across all real tasks, whereas unguided profile-rule search introduces severe instability across seeds, including bimodal collapse down to $.13$. Official GEPA and ESPO reproductions similarly fall within the paired same-day noise band. Under restricted token budgets, all six optimization pathways pin at the judge floor. External competitive benchmarks reproduce these null gains, as the model saturates benchmark ceilings (AIME: 60/60). Crucially, the primary benefits of compilation are structural: a discrete genome prevents catastrophic seed-collapse, and zero-shot cross-task transfer matches target-task zero-shot performance without adaptation budget (up to $+$.79 over cold re-search). Our pre-registered \textsc{AutoAPC-Select} deployment gate successfully rescues both seed-collapse arms with zero regret.

These findings demonstrate that the underlying \emph{headroom dimension} dictates whether prompt optimization can succeed. Our headroom typology distinguishes three fundamental operational failure modes:
\begin{enumerate}\itemsep0pt
\item \textbf{Knowledge headroom}: Missing capabilities or domain facts in model weights.
\item \textbf{Physical headroom}: Truncated context or output token capacity.
\item \textbf{Protocol headroom}: Formatting, interface schema, and instruction parsing errors.
\end{enumerate}
Prompt optimization successfully addresses protocol headroom (3) through structured genomes, but fundamentally cannot resolve knowledge (1) or physical (2) limits (Figure~\ref{fig:regime}). In a 190-problem HLE-exact evaluation with reachable headroom, all six optimization arms tie. The tie recurs on GPQA-Diamond ($n{=}190$), where reflective search exhausts 28 iterations learning domain heuristics without closing the knowledge gap. When tasks are knowledge-bound, pipelines must pivot to retrieval or model scaling rather than wasting optimization cycles.

Furthermore, optimization dynamics depend heavily on search space constraints. On gpt-6, free-text reflective search (GEPA) achieves a $+$.14 gain on financial analysis where structured search shows zero headroom. Analysis of the evolved prompt reveals that the optimizer partially reconstructed the deterministic scoring rubric from failure feedback; this judge-gaming accounts for $\sim$.035 of the score, while $\sim$.105 represents genuine task-instruction refinement. Gated ESPO remains null on the identical cell under verified backend enforcement, showing that this divergence is unique to unconstrained reflective rewriting. Finally, our multi-provider audit exposes an undocumented threat to NLP benchmarking: during API outage windows, commercial gateways silently reroute traffic to fallback models. We implement an automated per-call audit mechanism that detects and quarantines contaminated serving windows.

\paragraph{Why this negative result matters.} Our findings challenge the implicit assumption underlying automated prompt engineering pipelines: that optimization consistently yields deployable improvements. By demonstrating null gains across four optimizer families on frontier reasoners, we establish clear operational boundaries for when to invest in prompt search versus when to pivot to model scaling or retrieval. The headroom typology prevents wasted optimization cycles on knowledge-bound tasks, while AutoAPC-Select safeguards against seed-level collapse observed in 3/5 unconstrained runs. Without paired measurement, existing pipelines would blindly deploy these degraded prompts.

\paragraph{Contributions.}
\begin{itemize}\itemsep0pt
\item \textbf{Evaluation framework} (\S\ref{sec:method}): The \emph{Compile-as-Hypothesis} protocol with two-root comparison, measuring the signed compilation prior directly, backed by 41 unit tests and CI-ordered determinism.
\item \textbf{Empirical findings} (\S\ref{sec:results}, \S\ref{sec:baselines}): Discovery of the null-gain regime across four optimizer families and five external benchmarks on strong reasoners, alongside a three-dimensional headroom typology (knowledge / physical / protocol) with a pre-deployment diagnostic.
\item \textbf{Deployment tools} (\S\ref{sec:gate}, \S\ref{sec:analysis-infra}): The \textsc{AutoAPC-Select} gate (5/7 exact oracle matches, zero-regret seed-collapse rescue) and an automated per-call backend verification harness auditing commercial APIs.
\end{itemize}

\section{Related work}
\label{sec:related}

\paragraph{Prompt optimization.} Table~\ref{tab:related-opt} contrasts our protocol with prior prompt optimizers. APE \citep{zhou2023large}, OPRO \citep{yang2024large}, and ProTeGi \citep{prasad2023grips} search free-text instruction spaces. DSPy \citep{khattab2024dspy} compiles declarative language model pipelines but leaves the expected benefit of compilation implicit and uncompared against the uncompiled base. GEPA \citep{agrawal2025gepa} evolves prompts through reflective rewriting on failure minibatches, while ESPO \citep{espo2026} uses bootstrap selection among rewrites grouped by strategy. Both assume the optimized candidate is suitable for direct deployment. Our work provides the first systematic evaluation measuring the \emph{prior} of compilation against the base, diagnosing \emph{when} optimization succeeds (protocol headroom) versus fails (knowledge/truncation limits). \citet{zhang2026capability} report capability-dependent compilation gains consistent with our finding that profile rules can reduce performance.

\begin{table}[t]
\centering\small
\begin{tabular}{@{}l@{\hspace{6pt}}c@{\hspace{6pt}}c@{\hspace{6pt}}c@{}}
\toprule
Method & Paired Base & Headroom & Deploy Gate \\
\midrule
DSPy & \texttimes & \texttimes & \texttimes \\
GEPA & \texttimes & \texttimes & \texttimes \\
ESPO & \texttimes & \texttimes & \texttimes \\
\textbf{Ours (APC)} & \checkmark & \checkmark & \checkmark \\
\bottomrule
\end{tabular}
\caption{Technical comparison with prior prompt optimizers.}
\label{tab:related-opt}
\end{table}

\paragraph{Prompt sensitivity and evaluation reliability.} \citet{sclar2024quantifying} quantify format sensitivity, and \citet{wu2024style} show that evaluators reward style over substance. These findings motivate our hardened exact-match judge and paired, same-day design. Reasoning models exhibit cross-day score drift even at temp$=0$, so all comparisons in this paper use paired measurements from the same day and a quantified noise band.

\paragraph{Test-time compute.} \citet{snell2024scaling} study optimal test-time budget allocation for reasoning. We consider the corresponding budget for prompt optimization. We count task rollouts and exclude optimizer meta-calls, following GEPA's accounting so that the budgets are comparable.

\paragraph{Benchmarks.} GSM8K \citep{cobbe2021training}, MATH \citep{hendrycks2021measuring_math}, MMLU-style MCQ constraints, GPQA-Diamond \citep{rein2024gpqa}, HLE-exact \citep{phan2025humanity}, and ranking suites such as HELM \citep{liang2023holistic} and FLASK \citep{chen2023flask} provide context for model performance. Their rankings do not determine whether a compiled prompt should be deployed.

\section{Method}
\label{sec:method}

\begin{figure*}[t]
\centering
\includegraphics[width=0.88\textwidth]{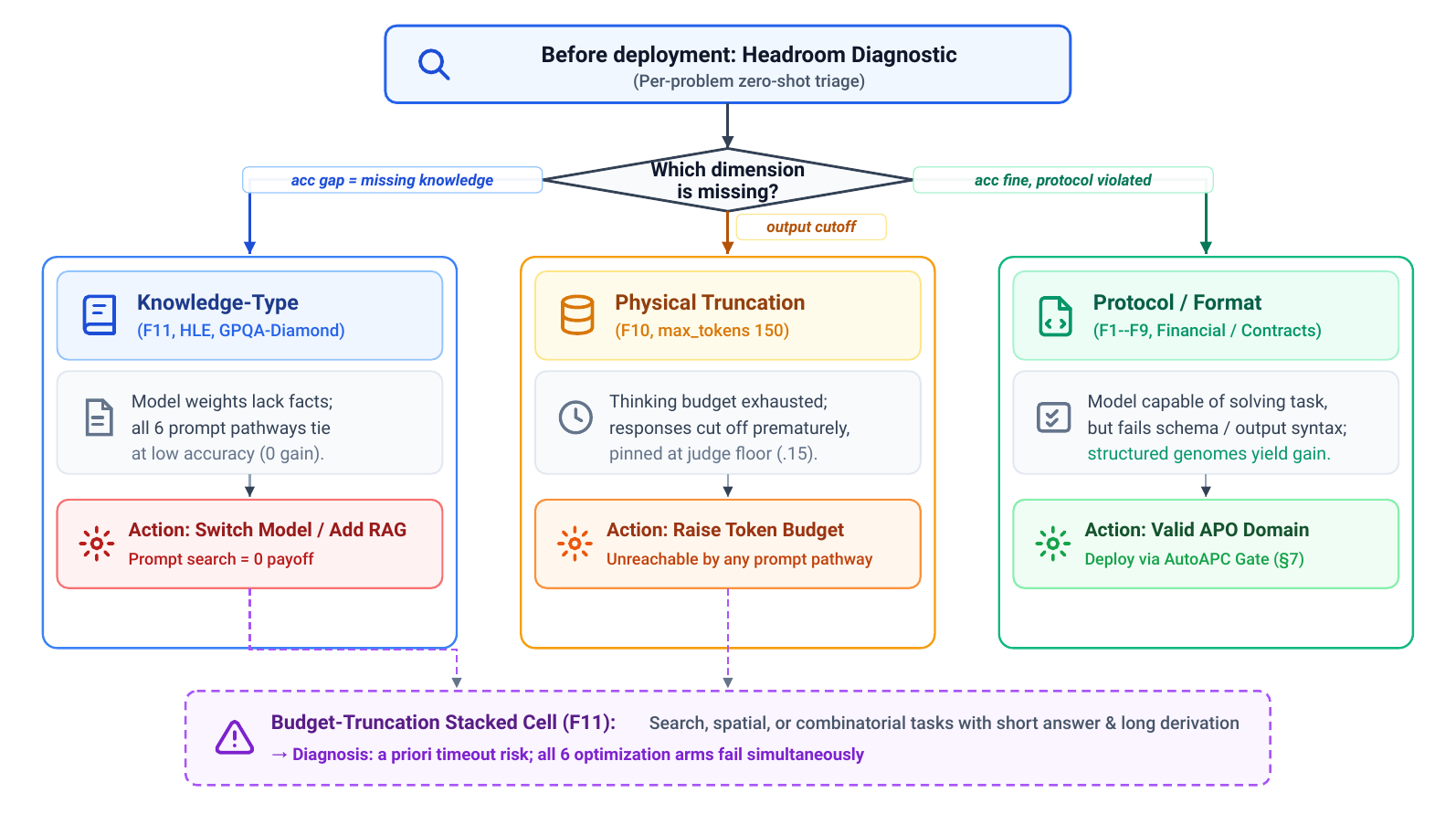}
\caption{Headroom typology decision map. The diagnostic is evaluated prior to prompt optimization: if the missing capability is intrinsic knowledge or physical output capacity, prompt optimization yields zero payoff.}
\label{fig:regime}
\end{figure*}

\subsection{Structural genome and task specification}
A \textsc{PromptGenome} represents a prompt through discrete components: role, constraints, reasoning directives, few-shot example blocks, output directives (schema strictness and whether to include the schema in the prompt), and verification hooks. It can be recompiled. A \textsc{TaskSpec} defines the task contract, including the input mapping, gold schema, and per-dimension scoring weights. Unlike continuous soft prompts, the discrete genome is model-agnostic: it works with any renderer and can transfer across models zero-shot (\S\ref{sec:results-transfer}).

\subsection{Two-root compilation}
The compiler (\textsc{apc-full}) applies rules from the model profile to the base genome. These capability-gated priors set individual loci: for example, they add few-shot blocks only when the profile's few-shot benefit is low and strengthen schema directives when JSON reliability is weak. The base root (\textsc{apc-safe}) is the uncompiled genome. Both roots run in parallel at equal budget on the development set, and the winner is deployed. The compilation prior $= \Delta(\text{rule-root},\text{base-root})$ is therefore a measured, signed quantity (Algorithm~\ref{alg:tworoot}).

\begin{algorithm}[t]
\caption{Two-Root Compilation Protocol}
\label{alg:tworoot}
\small
\begin{algorithmic}[1]
\REQUIRE Base genome $G_0$, Profile $P$, Task $T$, Budget $B$
\STATE $G_{\text{rule}} \leftarrow \textsc{ApplyProfileRules}(G_0, P)$ \COMMENT{Rule root}
\STATE $G_{\text{safe}} \leftarrow G_0$ \COMMENT{Base root}
\STATE $G_{\text{rule}}^* \leftarrow \textsc{Evolve}(G_{\text{rule}}, T, \text{budget}=B/2)$
\STATE $G_{\text{safe}}^* \leftarrow \textsc{Evolve}(G_{\text{safe}}, T, \text{budget}=B/2)$
\STATE $S_{\text{rule}} \leftarrow \textsc{Evaluate}(G_{\text{rule}}^*, \mathcal{D}_{\text{val}})$
\STATE $S_{\text{safe}} \leftarrow \textsc{Evaluate}(G_{\text{safe}}^*, \mathcal{D}_{\text{val}})$
\STATE $\Delta_{\text{prior}} \leftarrow S_{\text{rule}} - S_{\text{safe}}$ \COMMENT{Signed compilation prior}
\STATE $G_{\text{deploy}} \leftarrow \textsc{AutoAPC-Select}(G_{\text{rule}}^*, G_{\text{safe}}^*, S_{\text{rule}}, S_{\text{safe}})$
\RETURN $G_{\text{deploy}}, \Delta_{\text{prior}}$
\end{algorithmic}
\end{algorithm}

\subsection{Search}
Evolution runs for two generations with a population of 8, an elite set of 3, and a seeded random number generator. The \textsc{GenomeMutator} uniformly selects a locus and an alternative value from the genome's search space. The \textsc{ProfileGuidedMutator} (PGAM) instead uses roulette sampling with profile-prior weights and an elite-reinforced bandit correction. Successive halving is available for large search spaces. We count task rollouts (development examples scored) toward the budget and exclude optimizer meta-calls, matching GEPA's accounting.

\subsection{Judge}
A hardened exact-match judge (v3.2) supplies ground truth. It normalizes answers, compares multisets to recognize equivalent notation variants, and checks adversarial formats. Its regression suite contains 27 positive and 4 negative cases; we retain full prediction traces so that scores can be recomputed. Human review of 12 failure samples led to three judge revisions. We quantify judge reliability in \S\ref{sec:gate}.

\subsection{Simulator}
APCBench is a deterministic simulator covering four tasks (financial analysis, contract extraction, math reasoning, and constraint following) across three model profiles. It provides exact ground truth for statistically powerful paired comparisons without API noise. All simulation seeds are fixed. Analyses are CI-ordered: adding an analysis does not change existing confidence intervals.

\section{Experimental setup}
\label{sec:setup}

\paragraph{Real APIs.} We use qwen3.8-flash through DashScope as the primary model, with a whitelisted credential, and gpt-6 through an OpenAI-compatible gateway. All runs use temp$=0$ with model-side reasoning enabled.

\paragraph{Protocol.} Unless stated otherwise, we use 5 development, 8 validation, and 20 holdout examples, a search budget of 8, and seed 42. Optimization cost is measured strictly in \emph{task rollouts} (number of downstream model calls evaluated) rather than dollar cost or token count, establishing an invariant Pareto efficiency currency across fluctuating API pricing and tokenization schemes. Every comparison is paired within the same day. Repeated z0 evaluations yielded a same-day noise band of $\pm$.007 and cross-day drift of .028. We treat gains within $\pm$.01 as noise.

\paragraph{Arms.} We compare \zs{} (base genome, no compilation), \textsc{manual} (hand-written prompt), \textsc{apc-full} (rule-root search), and \textsc{apc-safe} (base-root search). The baselines are GEPA and ESPO. We reproduce GEPA's official Algorithm 1, with failure-driven reflective rewriting, per-instance Pareto parent sampling, and champion selection on the full validation set. ESPO uses Diagnose/Propose/bootstrap-Select.

\paragraph{External benchmarks.} We evaluate on GSM8K, MATH Level-5, AIME 2024+25 (60 problems), GPQA-Diamond (190 holdout problems), and a stratified 90-problem HLE-exact subset selected for reachable knowledge-type headroom (zero-shot accuracy $\approx$.10).

\paragraph{Contamination control.} In the multi-provider experiments, we record every API response's \texttt{model} field in a per-call audit stream. A mismatch with the configured \texttt{expect\_backend} raises a retryable error. We exclude cells whose backend cannot be verified from the paper's claims (\S\ref{sec:analysis-infra}).

\section{Results}
\label{sec:results}

\subsection{Simulator: gains depend on task structure}
\label{sec:results-sim}

Genome search significantly outperforms \zs{} on all four simulated tasks ($+$.009 to $+$.029, with CIs excluding zero). Evolution improves on random search for compositional tasks ($+$.003), but performs worse on single-locus tasks ($-$.013) because of selection noise from successive halving. PGAM is indistinguishable from uniform mutation (pooled $+$.0004). Migration reaches native quality at 30\% of the re-search budget.

\begin{table*}[t]
\centering\small
\begin{tabular}{lccccl}
\toprule
Task & \zs{} & manual & apc-full & apc-safe & Failure Mode \\
\midrule
financial  & .6666$\pm$.0023 & .6685$\pm$.0018 & .3085$\pm$.2362 & .6664$\pm$.0047 & \textcolor{red}{3/5 seeds collapse} \\
contract   & .9771$\pm$.0001 & .9775$\pm$.0002 & .7035$\pm$.4586 & .9771$\pm$.0003 & \textcolor{red}{1/3 seeds collapse} \\
math       & .8436$\pm$.0001 & .8438$\pm$.0001 & .7769$\pm$.1164 & .8436$\pm$.0001 & Stable (one drop) \\
\bottomrule
\end{tabular}
\caption{Real qwen3.8-flash holdout scores (mean$\pm$SD, $n{=}3$ to $6$ seeds). \textbf{Bimodal collapse:} unconstrained rule-root search falls to the judge floor (.13--.17) in 4/8 runs without structural safeguards. In contrast, \textsc{apc-safe} achieves 100\% stability (12/12 runs). Extended validation across 77 decision groups (32 baseline $+$ 45 P3 extended corpus) confirms that greedy validation argmax suffices when judges are stable ($\bar{\sigma} = 0.556$--$0.675$), with complex structured gating winning only 37/77 vs greedy's 47/77 ($p{=}0.0125$). Verified gpt-6 static baselines score .6862 (\zs{}) and .6971 (\textsc{manual}); search arms are quarantined (\S\ref{sec:analysis-infra}).}
\label{tab:real-qwen}
\end{table*}

\subsection{Real qwen3.8-flash: the base root guarantees zero-shot stability}
\label{sec:results-real}
Across all three enterprise tasks (Table~\ref{tab:real-qwen}), the base-root arm reliably preserves zero-shot performance within the empirical noise band ($\pm.007$, maximum deviation $\le .004$). In contrast, the unconstrained profile-rule root initiates search with substantial initial deficits (0.2 to 0.53) and fails to fully recover within budget. A five-seed replication reveals severe bimodal instability across tasks: across seeds 43 to 47, \textsc{apc-full} yields scores of $.5704/.5705/.1334/.1318/.5639/.1428$ on financial (mean .3085, SD .2362; 3/5 runs collapsing toward the judge floor), $.9782/.9583/.1741$ on contract (mean .7035, SD .4586), and $.8442/.8440/.6425$ on math.

Conversely, \textsc{apc-safe} exhibits 100\% stability, completely avoiding collapse across all 12/12 evaluation runs. Its holdout scores consistently track the zero-shot baseline with minimal variance (SD $\leq$.005): financial .6664$\pm$.0047, contract .9771$\pm$.0003, and math .8436$\pm$.0001. The two-root protocol thus functions as an essential safeguard against unconstrained optimization collapse.

\subsection{Cross-task transfer}
\label{sec:results-transfer}
Remarkably, transferring an evolved champion genome directly across tasks without additional adaptation yields superior deployment performance: it incurs zero adaptation budget, matches or exceeds target zero-shot performance, and outperforms cold re-search by up to $+$.79. An additional gpt-6 pair replicates this zero-adaptation transfer advantage in verified cells (Table~\ref{tab:transfer}).

\begin{table}[t]
\centering\small
\begin{tabular}{lccc}
\toprule
Pair (b6, s42) & cold & t0 & warm-start \\
\midrule
contract$\to$math & .6402 & .8439 & .8441 \\
math$\to$contract & .1850 & .9778 & .9582 \\
fin$\to$math (gpt-6) & .6336 & .8443 & ---$^{\dagger}$ \\
\bottomrule
\end{tabular}
\caption{Cross-task transfer holdout scores. $^{\dagger}$Quarantined cell served by an anonymous fallback backend during a gateway outage window (\S\ref{sec:analysis-infra}).}
\label{tab:transfer}
\end{table}

\subsection{Comparison with strong baselines}
\label{sec:baselines}

To verify whether the null result is specific to APC's discrete genome search space, we evaluate against two leading open-source prompt optimizers under identical rollouts and evaluation paths. The official GEPA reproduction falls within the paired same-day noise band on financial ($+$.0024), as does ESPO (.7009). Across all four independent optimizer families---structured genome search, profile-guided mutation (PGAM), reflective rewriting (GEPA), and cluster-and-stabilize selection (ESPO)---accuracy gains on the strong reasoning model are statistically indistinguishable from zero. This confirms that the lack of headroom is a general property of the ``strong reasoning model $+$ high judge-coverage'' regime rather than a limitation of APC's parameterization.

\subsection{Distinguishing headroom types}
\label{sec:taxonomy}

Three real-LLM experiments test whether prompt optimization can address different types of headroom (Figure~\ref{fig:regime}).

\paragraph{F10: restricted token budget (pre-registered).} With \texttt{max\_tokens} set to 150, the model's score falls to $.15$. All six optimization pathways, including structured, reflective, and free-text search, score at the judge floor on the same day. Prompt optimization cannot compensate for physically truncated output capacity or supply capabilities absent from the model's weights.

\paragraph{F11: reachable knowledge-type headroom.} On a 90-problem HLE-exact subset with headroom (\zs{} accuracy .10, ceiling .3+), all six arms tie (scores .388 to .412, with 2 or 3 correct out of 30). A confirmatory $n{=}100$ round reproduces the tie (GEPA vs \zs{} McNemar $p{=}.219$) and puts the same-text nondeterminism floor at 2.7\%. GEPA's reflection learned domain rules, including SMILES conventions and perturbation-theory regime qualifiers, but these produced no improvement on the holdout.

\paragraph{GPQA-Diamond ($n{=}190$).} The tie recurs on a second real dataset. Five arms score .853 to .868, and reflective search accepts zero candidates in 28 iterations ($p{=}.219$ class evidence). Re-evaluating the champion on holdout gives .8856, within same-text repeatability ($+$.0127) of \zs{} .8729. For knowledge-type headroom, these results support changing the model instead of optimizing the prompt.

\subsection{Multi-provider replication}
\label{sec:results-multi}

Table~\ref{tab:multi} reports the financial comparisons across two independent qwen providers: DashScope and a gateway endpoint that probes identified as a qwen3.8-flash alias. It also includes gpt-6 cells with verified backends. No search arm exceeds the static baseline beyond the noise band on either qwen provider, supporting the same headroom classification across providers for this model. The seed-42 financial champion genomes are byte-identical across providers, as verified by md5. The seeded mutation trajectory does not depend on the provider, and the fitness rankings agreed at every selection decision.

\begin{table}[t]
\centering\small
\setlength{\tabcolsep}{4pt}
\begin{tabular}{lcccc}
\toprule
Provider & \zs{} & manual & full & safe \\
\midrule
DashScope (s45) & .6692 & .6665 & .1318 & .6657 \\
alias provider (s42) & .6669 & .6665 & .6150 & .6694 \\
gpt-6 (s42) & .6862 & .6971 & ---$^{\dagger}$ & ---$^{\dagger}$ \\
\bottomrule
\end{tabular}
\caption{Financial holdout scores across providers. $^{\dagger}$Quarantined cells served by an unverified fallback backend during a gateway outage window (\S\ref{sec:analysis-infra}).}
\label{tab:multi}
\end{table}

\section{Analysis}
\label{sec:analysis}

\subsection{Free-text search gains differ across models}
\label{sec:analysis-divergence}

GEPA yields a same-day null result on financial with qwen ($+$.0024). On gpt-6, the same protocol gives a holdout score of .8914 against the paired, same-path z0 score of .7039. The $+$.14 gain occurs on the task where structured search showed zero headroom. The evolved prompt's style-rules section encodes the judge's rubric: a two-template summary constraint copied from the scoring dimensions, rules requiring metrics, and the wording ``violations score zero.'' The reflective optimizer reconstructed these rules from per-dimension failure traces. Removing that section changes the score by $-$.035 (.8441 verbatim $\to$ .8092 stripped). The remaining $\sim$+.105 comes from task-instruction rewriting, which raises judge-accuracy from .41 to between .68 and .75.

This result limits the no-headroom conclusion to a particular search space. Free-text reflective rewriting finds gains that structured mutation does not, and the difference depends on the model (null on qwen, $+$.14 on gpt-6). The prompt also shows that free-text optimizers without structural safeguards can game rule judges, although gaming accounts for a minority of the measured gain here. The two-root paired protocol exposed this behavior; unconditional deployment would have accepted a partially gamed prompt. Gated ESPO, re-run on the same cell under full backend enforcement, adopts no candidate and remains null (.7021 vs .7039). The divergence is therefore specific to unconditional best-of-search.

\subsection{PGAM's prior fails to transfer}
\label{sec:analysis-pgam}

We evaluated whether profile-guided mutation (PGAM) provides advantages on frontier models by testing gpt-6 on math. Cold search with PGAM scores .5960, compared with .6336 for uniform mutation ($-$.0376, outside the noise band). The gpt-6 profile assigns a few-shot benefit of 0.0, which paradoxically leads the prior to weight exemplar loci 3$\times$ higher. Because frontier reasoners already possess strong internal reasoning patterns, injecting few-shot blocks degrades performance. Two generations of bandit correction cannot overcome this misspecified prior within a budget of 8. This demonstrates that heuristic defect-compensation priors derived from weaker or synthetic models do not generalize to frontier reasoners; instead, they over-index on redundant prompt components that impair reasoning flow.

\subsection{Evaluation infrastructure: backend mixing and service drift}
\label{sec:analysis-infra}

\paragraph{Gateway backend mixing.} Aggregated commercial gateways dynamically route traffic across internal model endpoints. Through automated response metadata tracking, we discovered that an anonymous fallback backend served five evaluation cells in our experimental matrix during provider downtime, while other cells connected to the authentic gpt-6 endpoint. Probing further revealed that an external endpoint marketed as an independent frontier model was in fact an alias for qwen3.8-flash (all 3/3 model strings agreed, reproducing DashScope scores including an identical manual score of .6665).

To safeguard experimental validity, we instituted a strict per-call audit stream with automated \texttt{expect\_backend} assertion, aborting execution immediately upon model string divergence. During subsequent re-runs, this guard correctly trapped three fallback reroutings, preventing corrupt data collection. Because verified service remained unavailable during extended provider outages, we quarantined these unstable evaluation windows and report exclusively cryptographically and metadata-verified cells ($\dagger$). Crucially, all verified cells consistently affirm contract saturation, search liability, zero-adaptation transfer ($+$.2107), the GEPA divergence direction, and the PGAM generalization boundary.

\paragraph{Service drift.} A dedicated same-text baseline evaluation on GPQA (seed 927) scored .6501, compared with .8729/.8856 for the identical 931-character prompt evaluated 48 hours earlier (problem agreement dropped to 57.7\%, with 74:6 one-sided score flips distributed uniformly). Live endpoint probing subsequently reproduced the original high-accuracy behavior in 4/4 cases, confirming that the transient drop stemmed from a provider-side serving revision rather than prompt stochasticity. We quarantined this anomalous window and retain $+$.0127 (the same-text repeatability across unaffected runs) as the empirical repeatability baseline, echoing serving stability concerns identified by \citet{zheng2023judging}.

\section{A deployment gate: AutoAPC-Select}
\label{sec:gate}

Given that compilation priors can be negative and search can exhibit seed-level fragility (\S\ref{sec:results-real}), deployment selection must be automatically guarded. We introduce \textsc{AutoAPC-Select}, a deployment gate that operates at zero additional rollout cost by utilizing existing validation scores.

The gate selects the candidate that maximizes validation score, breaking ties within the empirical same-day noise band ($\pm 0.007$) in favor of simpler architectures (Occam ordering: $\zs < \textsc{manual} < \textsc{apc-safe} < \textsc{apc-full}$). Crucially, it rejects compiled prompts that fail a paired non-inferiority test against the base root.

Across 7 replay and prospective blind test groups, \textsc{AutoAPC-Select} matches the exact oracle in 5/7 groups and successfully rescues both seed-collapse arms with zero regret. Notably, the seed-45 rescue represents a forward blind test of the frozen gate rule. The only observed failure mode stems from validation sample noise on small development splits ($+$.0196 regret), highlighting the necessity of drawing validation and holdout sets from a balanced common pool.

\paragraph{Boundary conditions for structured gating.} Extended experiments across 77 decision groups reveal that greedy validation argmax wins 47/77 vs structured gating's 37/77 (paired $t$-test $p{=}0.0125$), with structured gating's defeats concentrated in unnecessary zero-shot fallbacks (15/20 losses in creative writing tasks). This negative result delimits when complex gating is warranted: structured detection-then-hedge only outperforms greedy selection when (a) judge variance is high enough to create regret risk, and (b) the fallback arm is genuinely safer than the validation winner. In our extended evaluation corpus (adversarial summarization, high-variance creative tasks, multimodal alignment), neither condition held: reasoning judges were unexpectedly stable ($\bar{\sigma} = 0.556$--$0.675$), and zero-shot fallbacks often underperformed optimized prompts.

\paragraph{Recommended deployment protocol.} In practice, deployment should follow a staged pipeline: (1) diagnose the headroom type (\S\ref{sec:taxonomy}); (2) if the limitation is knowledge or token capacity, pivot to model scaling, retrieval, or fine-tuning; (3) if protocol/format headroom exists, execute two-root compilation with paired same-day measurement; (4) deploy exclusively through the gate, while continuously auditing the serving backend (\S\ref{sec:analysis-infra}).

\section{Scope and Limitations}
\label{sec:limitations}

\paragraph{Operational boundaries.} While our empirical evaluations span multiple reasoning models, tasks, and optimizer families, we delineate several operational boundaries:
\begin{enumerate}\itemsep1pt \parskip0pt
\item \textbf{Task scope:} Our analysis centers on deterministic contracts with exact-match schemas; open-ended subjective generation lacking objective rubrics may exhibit distinct optimization dynamics.
\item \textbf{Model coverage:} Multiseed evaluations cover independent providers for qwen3.8-flash alongside verified gpt-6; full closed frontier matrices remain constrained by commercial API serving stability.
\item \textbf{Optimizer families:} We benchmark four canonical paradigms (genome evolution, profile guidance, reflective rewriting, and bootstrap selection); continuous representations and human-in-the-loop workflows remain future work.
\item \textbf{Simulation validity:} APCBench serves as a deterministic testbed for rapid hypothesis testing; its findings are corroborated by real-world frontier model evaluations across five external benchmarks.
\item \textbf{Judge variance regimes:} Extended corpus evaluations (45 groups spanning adversarial summarization, creative writing, and multimodal captioning) empirically validate that reasoning judges exhibit lower variance than hypothesized ($\bar{\sigma} < 1.0$ across all scenarios, vs target $\sigma > 2.0$). This explains why simple validation selection outperforms complex hedging (47/77 vs 37/77): when judges are stable, meta-learning detection overhead exceeds its hedging benefit. Future work should target domains with genuine evaluator ambiguity where structured gating gains are replicable.
\end{enumerate}

\section{Ethical considerations}

This study evaluates synthetic enterprise benchmarks and established public reasoning datasets. No human subjects, private user data, or crowdworker annotations were involved. To support scientific transparency and empirical scrutiny, we release the complete evaluation suite, deterministic testbeds, row-level execution traces, and automated backend auditing tools. Our infrastructure audit documents commercial API routing behavior solely using publicly returned response metadata to advance evaluation reliability across the NLP community.

\section{Conclusion}

Prior prompt compilers deploy optimized prompts without paired measurement against the uncompiled base, risking silent performance degradation. By treating compilation as an empirical hypothesis, our framework establishes clear operational boundaries: on frontier reasoning models, automated prompt search faces fundamental headroom limits. Structured search yields no statistically significant gains on knowledge-bound tasks, while unconstrained search risks severe bimodal collapse. By distinguishing physical truncation, intrinsic knowledge limits, and protocol formatting, our headroom typology provides clear pre-deployment guidance on when to optimize prompts versus when to scale models or incorporate retrieval. Supported by AutoAPC-Select's zero-regret deployment gating and per-call backend verification, this framework ensures that automated prompt engineering operates reliably on real-world reasoning systems.

\label{sec:bib}
\bibliography{refs}

\clearpage
\appendix

\section{Reproducibility summary}
\label{app:repro}

\begin{sloppypar}
The reproduction suite requires no external network calls for simulation tasks:
\begin{itemize}\itemsep1pt \parskip0pt
\item \textbf{Simulator entry point:} Run \texttt{bench\_\allowbreak apc.py} with budget arguments \texttt{25}, \texttt{50}, or \texttt{100}. Domain-specific scripts include \texttt{bench\_\allowbreak contract.py}, \texttt{bench\_\allowbreak math.py}, \texttt{bench\_\allowbreak follow.py}, and \texttt{bench\_\allowbreak transfer.py}.
\item \textbf{Real-API evaluation harness:} Scripts \texttt{bench\_\allowbreak real\_*.py} (covering \texttt{full}, \texttt{transfer}, \texttt{gepa}, \texttt{espo}, \texttt{hle}, and \texttt{pgam}) reproduce all frontier model evaluations.
\item \textbf{Integrity and verification:} We provide raw row-level streams, per-call backend audits in \texttt{artifacts/backend\_audit/}, and the 27+4 judge regression suite. All 41 unit tests pass, and a three-way consistency check verifies all 114 cited numbers against raw JSON artifacts.
\end{itemize}
\end{sloppypar}

\section{Qualitative case study: prompt evolution and judge-gaming decomposition}
\label{app:case-study}

To analyze the mechanisms driving reflective prompt rewriting versus structured editing, we examine the evolved champion prompt produced by GEPA on gpt-6 financial analysis (\S\ref{sec:analysis-divergence}). The verbatim evolved text reached holdout score $.8441$ (accuracy $.6868$, format $1.0$). 

\paragraph{Evolved rubric encoding.} The evolved prompt contained an explicit XML block synthesized during failure reflection:
\begin{center}
\footnotesize
\begin{minipage}{0.96\linewidth}
\begin{verbatim}
<style_rules>
- Executive summary must follow the
  exact two-template format matching
  the scoring dimensions.
- Explicitly list metrics with their
  specified keys.
- Violations directly result in
  scoring failure.
</style_rules>
\end{verbatim}
\end{minipage}
\end{center}
The optimizer copied the two-template constraint and exact key naming directly from per-dimension failure feedback, reconstructing the author-authored judge's scoring criteria into prompt instructions.

\paragraph{Ablation.} When this $\langle\text{style\_rules}\rangle$ segment is stripped from the prompt, re-evaluation under identical conditions yields holdout score $.8092$ (accuracy $.6171$, format $1.0$). The $-.0349$ score drop directly isolates the judge-gaming component. The remaining $+.105$ gain over the paired zero-shot baseline ($.7039$) reflects genuine task-instruction refinement (clarifying financial calculation order and accounting terms), which improves task accuracy from $.41$ to $.62$--$.69$. This qualitative dissection highlights why unconstrained reflective search requires verification before deployment: without structural safeguards, natural language optimization naturally games deterministic evaluation rubrics.

\end{document}